\documentclass[runningheads]{llncs}
\usepackage{graphicx}
\usepackage{amsmath,amssymb} 
\usepackage{color}
\usepackage{rotating}
\usepackage{url}

\begin{document}
\pagestyle{headings}
\mainmatter

\def\ACCV16SubNumber{17}  

\title{STFCN: Spatio-Temporal
	FCN\\ 
	for Semantic Video Segmentation} 
\author{Mohsen Fayyaz\textsuperscript{1}, Mohammad Hajizadeh Saffar\textsuperscript{1}, Mohammad Sabokrou\textsuperscript{1}, Mahmood Fathy\textsuperscript{2}, Reinhard Klette\textsuperscript{3}, Fay Huang\textsuperscript{4}}
\institute{\textsuperscript{1}Malek-Ashtar University of Technology, \textsuperscript{2} Iran University of Science and Technology, \textsuperscript{3}Auckland University of Technology, \textsuperscript{3}National Ilan University}
\maketitle

\begin{abstract}
This paper presents a novel method to involve both spatial and temporal 
features for semantic segmentation of street scenes. Current work on {\it convolutional neural 
networks} (CNNs) has shown that CNNs provide advanced spatial features
supporting a very good performance of solutions  for the semantic 
segmentation task. We investigate how involving temporal features also 
has a good effect on segmenting video data. We propose a module based 
on a {\it long short-term memory} (LSTM) architecture of a 
recurrent neural network for interpreting the temporal characteristics of 
video frames over time. Our system takes as input frames of a video and 
produces a correspondingly-sized output; for segmenting the video our 
method combines the use of three components: First, the regional spatial 
features of frames are extracted using a CNN; then, using LSTM the 
temporal features  are added; finally, by deconvolving the spatio-temporal 
features we produce pixel-wise predictions. Our key insight is to build 
{\it spatio-temporal convolutional networks} (spatio-temporal CNNs) that 
have an end-to-end architecture for semantic video segmentation. We 
adapted fully some known convolutional network architectures 
(such as FCN-AlexNet and FCN-VGG16), and dilated convolution into 
our spatio-temporal CNNs. Our spatio-temporal CNNs 
achieve state-of-the-art semantic segmentation, as demonstrated for 
the Camvid and NYUDv2 datasets.
\end{abstract}

\section{Introduction}

Semantic  segmentation of video data is a fundamental task  
for  scene understanding. For many computer vision applications, semantic 
segmentation is considered as being (just) a pre-processing task.  
Consequently, the performance of semantic segmentation has a 
direct effect on subsequent computer vision solutions which depend on it. 
Self-driving cars is one of the areas in technology that has received 
much attention recently.  
These cars can detect surroundings using advanced driver assistance systems (ADAS) 
that consist of many different systems such as radar, GPS, computer vision, 
and in-car networking to bring safety to driving and roads.
One of the main processes for the computer vision part of these systems 
can be identified as being semantic segmentation of all objects in 
surroundings to transmit accurate and complete information to 
the ADAS system such that the system can make the best decision 
to avoid accidents. 

Segmentation is typically approached as a classification problem. First, using a set 
of labeled video frames, the characteristics of all segments (classes) are learned. 
These characteristics are used for labeling the pixels of test frames \cite{BAD2015,YU2016}. 
Recently, deep learning methods, especially CNNs, ensured state-of-the-art performance 
in different areas of computer vision, such as in image 
classification~\cite{KRI2012}, object detection~\cite{GIR2014}, or 
activity recognition~\cite{SIM2014}.

We consider the application of advanced features, 
extracted by using CNNs, for semantic video segmentation. 
Semantic segmentation methods use both given image data at 
selected locations as well as a semantic context. A set of pixels 
is usually predicted as defining one class (or even one segment) 
if connected, and also referring to one particular semantic interpretation. 

Previous methods for video segmentation
have efficiently exploited CNNs, but they did not use 
temporal features; of course, temporal features can be useful for 
interpreting a video semantically. For example, 
the authors of \cite{BAD2015,YU2016} represented and interpreted 
video frames using a deep learning method, but the main disadvantage of 
their methods is that they consider those frames as being independent from 
each other. Neglecting the time dimension in video data basically
means that the given raw data are down-sampled without 
using fully given information. Using temporal features can help the 
system to distinguishing, for example, between two objects of 
different classes having the same spatial features but showing 
differences in the time feature dimension. 

Consequently, we propose a method which uses a similar 
paradigm for extracting spatial features (as in the cited papers), but which
differs by also using temporal features (i.e. features of a continues 
sequences of frames). We propose to identify components 
which can be embedded ``on top'' of spatially extracted features maps 
in individual frames. Such a component can be seen as being
equipped with a set of memory cells which save the assigned 
regions in previous frames. This allows us that relations between 
regions, available in previous frames, can be used to define 
temporal features.  We process the current video frame 
by using the spatio-temporal output features of our processing 
modules. 

\begin{figure*}[t]
\center
	\includegraphics[width=1\linewidth]{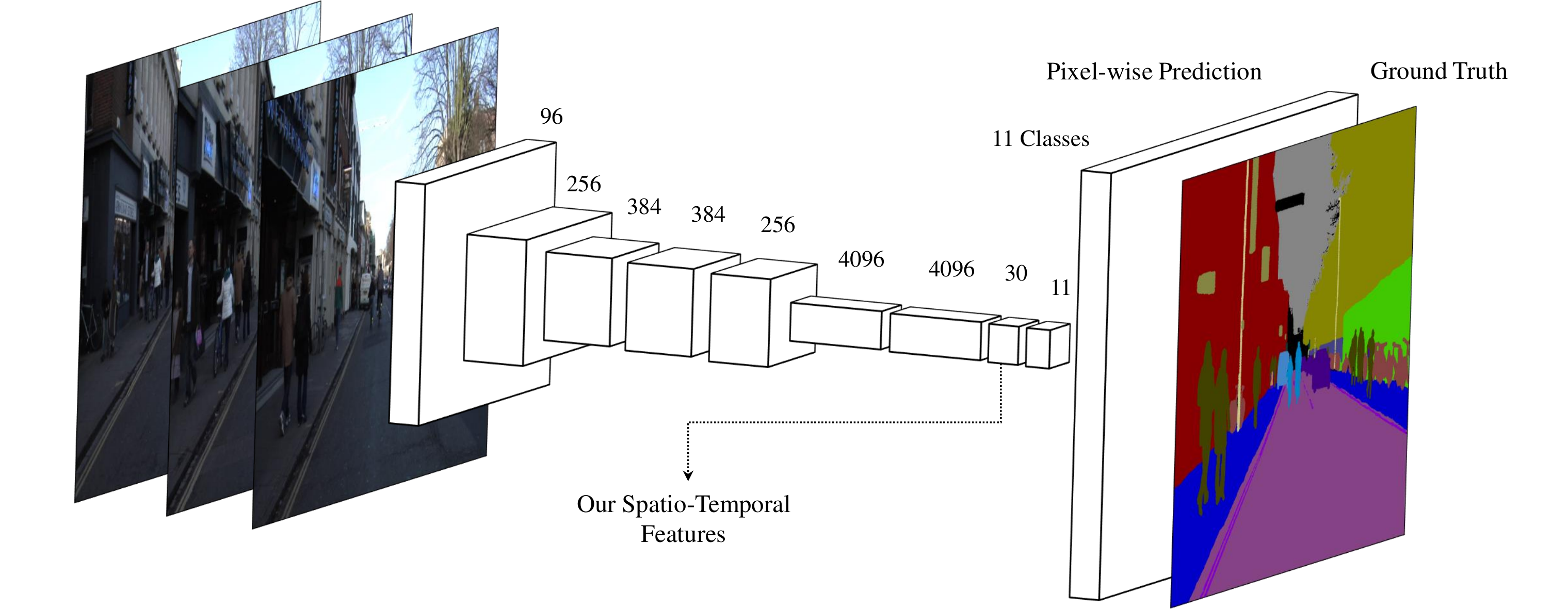}
	\label{fig:1}
	\caption
	{
		A spatio-temporal fully convolutional Alexnet architecture, later 
		also to be discussed in Section~\ref{sec:E_FCN}
	}
\end{figure*}

Similar to other segmentation methods, we use 
then some fully convolutional layers to perform regional semantic 
classification. In our method, these fully convolutional layers perform 
spatio-temporal classifications. Finally, we use a deconvolution 
procedure for mapping (i.e. scaling) the obtained predictions into the original 
carrier (i.e. the image grid) of the given frames for having a pixel-wise 
prediction. See Fig.~\ref{fig:1}.

CNN-based methods usually combine two components, where one is 
for describing and inferring a class of different regions of a video frame 
as a feature map, and another one for performing an up-sampling 
of the labeled feature maps to the size of the given video frames.  
An advantage of our method is that we can adjust and embed 
our proposed module into the end of the first component (before inferring 
the labels) of current CNN-based methods as an end-to-end network. 
We show that the proposed changes in the network lead to an 
improvement in the performance of state-of-the-art methods, such as, 
FCN-8 \cite{LON2015} and dilated convolution \cite{YU2016}.

The main contributions of this paper are as follows:
\begin{itemize}
\item The proposed method can be easily adapted for enhancing already published 
state-of-the art methods for improving their performance.
\item We propose an end-to-end network for semantic video segmentation 
in respect to both spatial and temporal features.
\item We propose a module for transforming traditional, fully convolutional 
networks into spatio-temporal CNNs.
\item We outperformed state-of-the art methods on two standard benchmarks. 
\end{itemize}
The rest of this paper is organized as follows. Top-ranked related work on 
semantic video segmentation is reviewed in Section~\ref{sec:rw}. 
Section~\ref{sec:pm} introduces the proposed method. The performance of 
our method is shown in Section~\ref{sec:er}. 
Section~\ref{sec:con} concludes the paper. 

\section{Related Work}
\label{sec:rw}

There is a wide range of approaches that have been published 
so far for video segmentation. Some of them have 
advantages over others. These approaches can be categorized 
based on the kind of data that they operate on, the method 
that is used to classify the segments, and the kind of segmentation 
that they can produce.

Some approaches focus on binary classes such as 
foreground and background 
segmentation \cite{BIT2015,CAR2010}. 
This field includes also some work that has a focus on 
anomaly detection \cite{SAB2015} since authors use a single-class classification 
scheme and constructed an outlier detection method for all other categories. 
Some other approaches concentrate on multi-class 
segmentation \cite{CHE2010,LIU2015,LIU2013,YAN2012}.

Recently created video datasets provide typically image data in RGB format.
Correspondingly, there is no recent research on gray-scale semantic video 
segmentation; the use of RGB data is common standard, 
see \cite{GAL2014,KHO2015,LIU2015,LIU2013,ZHA2015}. 
There are also some segmentation approaches 
that use RGB-D datasets \cite{HE2016,HIC2014,MAR2015}.

Feature selection is a challenging step in every machine learning approach. 
The system's accuracy is very much related to the set of features that are 
chosen for learning and model creation. Different methods have been proposed 
for the segmentation-related feature extraction phase.

\subsection{Feature Extraction}

We recall briefly some common local or global feature extraction 
methods in the semantic segmentation field. These feature extraction 
methods are commonly used after having super-voxels extracted from 
video frames \cite{LIU2013}.

Pixel color features are  
features used in almost every semantic segmentation 
system \cite{GAL2014,KHO2015,LIU2015,LIU2013,MAR2015}. 
Those includes three channel values for RGB or HSV images,
and also values obtained by histogram equalization methods. The 
{\it histogram of oriented gradients} (HOG) defines a set of features combining 
at sets of pixels approximated gradient values for 
partial derivatives in $x$ or $y$ direction\cite{KHO2015,LIU2015}. 
Some approaches also used other histogram definitions such as the hue 
color histogram or a texton histogram \cite{ZHA2015}. 

Further appearance-based features are defined as across-boundary  
appearance features, texture features, or spatio-temporal appearance 
features; see \cite{GAL2014,KHO2015,LIU2015,LIU2013}. 
Some approaches that use RGB-D datasets, also include 
3-dimensional (3D) positions or 
3D optical flow features \cite{HIC2014,MAR2015}. 
Recently, some approaches are published that use CNNs for feature extraction;
using pre-trained models for feature representation is 
common in  \cite{BAD2015,HE2016,ZHU2015}.

After collecting a set of features for learning, a model must be chosen for 
training a classifier for segmentation. Several methods 
have been provided already for this purpose, and we recall a few.

\subsection{Segmentation Methods}

Some researches wanted to propose a (very) general image 
segmentation approach. 
For this reason, they concentrated on using unsupervised segmentation. 
This field includes clustering algorithms such as k-means 
and mean-shift \cite{LIU2013a}, or graph-based 
algorithms \cite{GAL2014,HIC2014,KHO2015,ZHA2013}.

A {\it random decision forest} (RDF) can be used for
defining another segmentation method that is a kind of a classifier 
composed of multiple classifiers which are 
trained and enhanced by using randomness
extensively \cite{GUP2014,RIC2015}. 
The {\it support vector machine} (SVM) \cite{VOL2015} or
a {\it Markov random field} (MRF) \cite{SHA2015,ZHE2015} 
are further methods used for segmentation but not as popular as 
the {\it conditional random field} (CRF) that is in widespread use 
in recent work\cite{CHA2014,LIU2015,MOT2013}.

Neural networks are a very popular method for image segmentation, 
especially with the recent success of using convolutional neural network 
in the semantic segmentation field. Like for many other vision tasks, 
neural networks have become very 
useful \cite{BAD2015,GIR2014,HE2016,HON2015,LON2015,ZHU2015}.

{\it Fully convolutional networks} (FCNs) are one of the 
topics that interest researchers recently. An FCN is based on the idea 
of extending a {\it convolutional network} (ConvNet) 
for arbitrary-sized inputs \cite{LON2015}. On the way 
of its development, it has been used for 
1-dimensional (1D) and 2-dimensional (2D) inputs \cite{MAT1991,WOL1994}, 
and for solving various tasks such as image restoration, sliding window 
detection, depth estimation, boundary prediction, or semantic segmentation. 
In recent years, many approaches use ConvNets as feature 
extractor \cite{BAD2015,HE2016,ZHU2015}. Some approaches turn ConvNets into 
FCNs by discarding the final classifier layer, and convert all fully connected 
layers into convolutions. By this change, authors use 
a front-end module 
for solving their vision tasks 
\cite{BAD2015,GIR2014,HE2016,HON2015,LON2015,ZHU2015}.

Recently, a new convolutional network module has been introduced 
by Yu and Fisher \cite{YU2016} that is especially designed for dense 
prediction. It uses dilated convolutions for multi-scale contextual 
information aggregation, and achieves some enhancements in 
semantic segmentation compared to previous methods. 
Kundu and Abhijit \cite{KUN2016} optimized the mapping of 
pixels into a Euclidean feature space; they achieve even better results 
for semantic segmentation than \cite{YU2016} by using a graphical CRF model.

Many approaches that have been introduced in this field have 
not yet used temporal features, especially in the field of deep CNNs \cite{GAL2014,HIC2014,KHO2015,LIU2015,LIU2013,ZHA2015}. 
These approaches cannot be identified as being 
end-to-end methods, which points to an essential disadvantage 
when applying these approaches. Some 
approaches use deep CNNs \cite{HE2016,KUN2016} by 
introducing an end-to-end architecture 
for also using spatio-temporal features for semantic labeling. 
However, none of them can change the size of time windows 
dynamically. 

Long short-term memory (LSTM) is a memory 
cell module that was introduced by \cite{GER2000,HOC1997}. 
It has many advantages such as the ability to support very 
large time windows, the ability to change time windows dynamically, 
the ability to handle noise, distributed representations, 
continuous values, and so forth. We propose for the first time an approach 
that uses a deep CNN network with LSTM modules as an end-to-end 
trainable architecture for semantic video segmentation and labeling.

\section{The Proposed Method}
\label{sec:pm}

\begin{figure}[t]
	\center
	\includegraphics[width=1\linewidth]{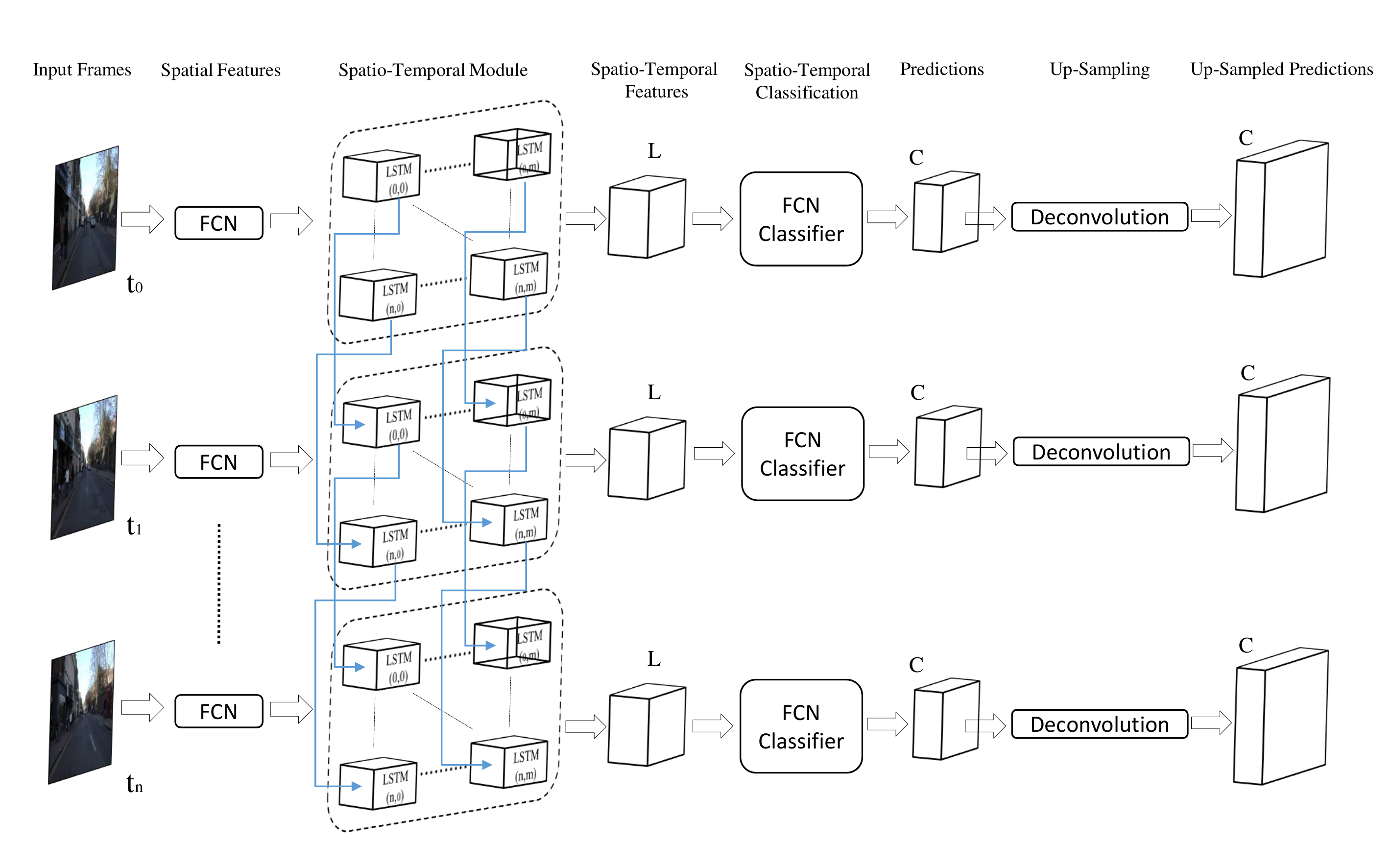}
	\caption
	{
		Overall scheme for our proposed end-to-end network architecture. 
		The LSTMs are used for inferring the relations between spatial features 
		which are extracted from the frames of the video
	}
	\label{fig:2}
	
\end{figure}

\subsection{Overall scheme} 
\label{sec:os}

We have four key steps in our method as shown in Fig. \ref{fig:2}. We 
feed the frame $I_t$ (i.e. the $t^{th}$ frame of a video), 
into a FCN network. 
This network down-samples the input images and describes a 
frame $I_t$, defined on an image grid $\Omega$ of size ${W \times H}$, 
as a features set $S_{t}^{1..m}$ in $m$ different maps.
The input is $I_t$ and the output of the 
latest layer (i.e. of lowest resolution) of the FCN 
is $S_{t}^{1..m}$ of size ${W' \times H'}$,
where $W' \ll W$ and $H' \ll H$. As a result, frame $I_t$ is represented 
as a  feature set $\{S_{t}^{1..m}\}$.
Every point $(i,j)$, with $1 \le i \le W'$ and $1 \le j \le H'$, in $S_{t}^{1..m}$ is 
a descriptor of size $m$ for a  region (receptive field) in $I_{t}$. 

We put our {\it spatio-temporal module} on top of the 
final convolutional layer. So, feature set $\{S_{t}^{1..m}\}$ will be represented 
as a spatio-temporal feature set of $\{ST_{t}^{1..m}\}^{(i,j)}$ by 
our {\it spatio-temporal module}. By applying an FCN classifier layer on top of 
these features, we predict the semantic classes of these regions in the video. 
Finally, we up-sample these predictions to the size of the $I_t$ frame. In following subsections, the methodologies that have been used in this approach, will be described.

\subsection{Fully Convolutional Network}

Convolutional neural networks (CNNs) are applied for a large set 
of vision tasks. Some researchers improve CNNs by changing its 
basic architecture and introducing new architectures. Recently, 
fully convolutional networks (FCNs) have been introduced by 
discarding the final classifier layer, and by converting all fully 
connected layers into convolutional layers. We follow this principle.

\subsection{LSTM}

\begin{figure*}[t]
	\center
	\includegraphics[width= 0.7\linewidth]{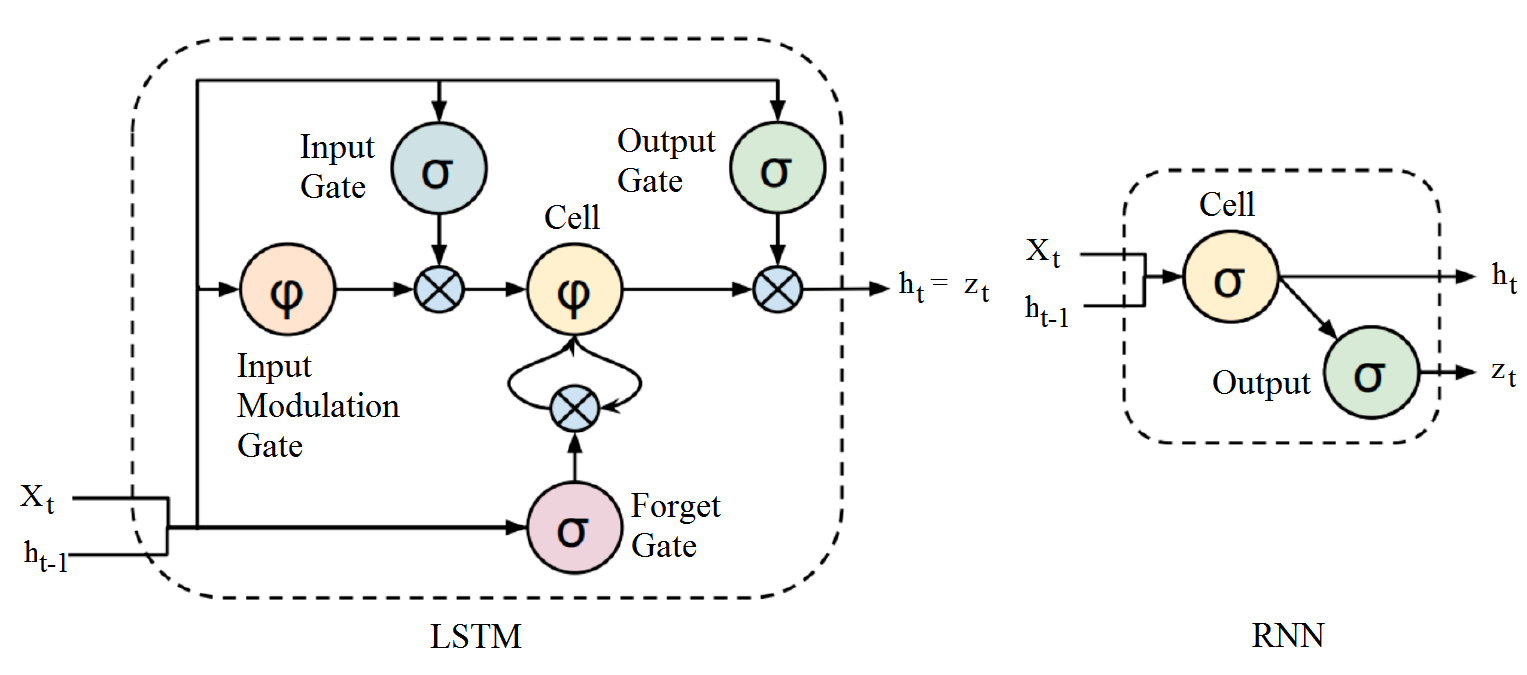}
	\caption
	{
		An example of a basic LSTM cell ({\it left}) and 
		a basic RNN cell ({\it right}). Figure follows a
		drawing in \cite{DON2015}
	}
	\label{fig:lstm}
\end{figure*}

A long short-term memory (LSTM) network is a special kind of 
{\it recurrent neural networks} (RNNs) that have been introduced 
by \cite{HOC1997} to solve the vanishing gradient problem 
and for remembering information over long periods. For 
an example of a basic RNN and an LSTM cell, see Fig.~\ref{fig:lstm}. 
LSTMs are not confined to fixed-length inputs or outputs, and this advantage 
makes them powerful for solving sequential problems.

Each LSTM module consists of a memory cell and a number of 
input and output gates that control the information flow in a sequence 
and prevent it from loosing important information in a time series. 
Assuming $S_t$ as the input of an LSTM module at time $t$, 
the cell activation is as formulated in the following equations:
\begin{eqnarray}
i_t&=&\sigma(W_{xi}x_t+W_{hi}h_{t-1}+b_i)
\label{eq:2}
\\
f_t&=&\sigma(W_{xf}x_t+W_{hf}h_{t-1}+b_f)
\label{eq:3}
\\
o_t&=&\sigma(W_{xo}x_t+W_{ho}h_{t-1}+b_o)
\label{eq:4}
\\
g_t&=&\phi(W_{xc}x_t+W_{hc}h_{t-1}+b_c)
\label{eq:5}
\\
c_t&=&f_t \odot c_{t-1} + i_t \odot g_t
\label{eq:6}
\\
h_t&=&o_t \odot \phi (c_t)
\label{eq:7}
\end{eqnarray}
where $\sigma$ and $\phi$ are symbols for 
a sigmoid and the ${\rm tanh}$ function, respectively. 
Symbol $h_t \in R^N$ denotes a hidden state 
with $N$ units, and $c_t \in R^N$ is the memory cell. 
By $i_t \in R^N$, $f_t \in R^N$, $o_t \in R^N$, and $g_t \in R^N$ we denote 
the input gate, forget gate, output gate, and input modulation gate at time $t$, 
respectively. Symbol $\odot$ stands for element-wise multiplication.

\subsection{Spatio-Temporal Module}

In regards to every $W' \times H'$ region 
of $I_{t}$, which is described by an FCN as an $\Omega$ grid, 
an LSTM is embedded (see Section~\ref{sec:os}). Thus we have 
altogether $W' \times H'$ 
LSTMs. Element $\{S_{t}^{1..m}\}^{(i,j)}$ defines a spatial 
characteristics of a region in the $I^t$ frame. These characteristics 
are given to LSTM$^{(i,j)}$ for processing; it infers a relation with
spatial features of equivalent regions in 
frames previous to frame $I_{t}$. With this ``trick'', 
both spatial and temporal features of a frame are considered.
(Note that LSTM$^{(i,j)}(\{S_{t}^{1..m}\}^{(i,j)})=\{ST_{t}^{1..m}\}^{(i,j)})$ where 
$S$ and $ST$ are spatial and 
spatio-temporal features, respectively). 

We embed one LSTM for each region. Equation~(\ref{eq:1}) shows a
representation of frame $I_t$ with respect to our suggested 
spatial and temporal features:
\begin{equation}
\Omega'_t(i,j)=(LSTM^{(i,j)}(\Omega_t(i,j))
\label{eq:1}
\end{equation}
where the size of $\Omega'$  is equal to that of $\Omega$, 
and value $m$ specifies a
map which assigns spatio-temporal features to every point for describing 
an equivalent region (i.e. a segment) in $I_t$. Now, similar to other 
methods \cite{BAD2015,LON2015,YU2016}, 
the labels for points in $\Omega'$ are predicted and up-sampled 
to the frame at the original size. The overall update function
can be briefly specified as follows:
\begin{equation}
\Omega'_t(i,j)=\sigma(W_{xo}x_t+W_{ho}h_{t-1}+b_o) \odot \phi (f_t \odot c_{t-1} + i_t \odot g_t)
\label{eq:8}
\end{equation}
Altogether, we introduced an operator layer to several LSTMs for 
properly representing the temporal features.  

This proposed network executes and processes the input frames 
as an end-to-end network.  
Figure~\ref{fig:2} shows the overall scheme of our method.

\subsection{Deconvolution}

Interpolation is a common method for mapping outputs into dense pixels. 
There are several interpolation (or upsampling) algorithms such as 
bilinear, non-linear, cubic, and so forth. Up-sampling by a factor $k$ 
can be considered as being a convolution with a fractional input 
stride of $1/k$. As a result, a convolution operator with input stride of 
$1/k$ can be applied backward (called deconvolution) with 
a stride of $k$ \cite{LON2015}.

\section{Experimental Results}
\label{sec:er}

For implementing our {\it spatio-temporal fully convolutional network} 
(STFCN) we use the standard Caffe distribution \cite{Jia2014} 
and a modified Caffe library with an LSTM implementation.\footnote
    {
    Available at \url{github.com/junhyukoh/caffe-lstm}
    } 
We merged this LSTM implementations into the 
Caffe standard distribution and released our modified 
Caffe distribution to support new FCN layers that have been
described in \cite{LON2015}. Our code has been tested on 
NVIDIA TITAN, and NVIDIA TITAN-X GPUs.\footnote
    {
    Our modified Caffe distribution and STFCN models 
    are publicly available at \url{https://github.com/MohsenFayyaz89/STFCN}.
    }

To show the performance of our modified version of 
FCNs we use their implemented models for two cases, 
with and without our spatio-temporal module. 
We tested our STFCN networks on Camvid\footnote
   { 
   Available at \url{mi.eng.cam.ac.uk/research/projects/VideoRec/CamVid/}
   } 
and NYUDv2\footnote
   {
   Available at \url{cs.nyu.edu/~silberman/datasets/nyu_depth_v2.html}
   } 
datasets. Our evaluation methodology is as in other state-of-the-art semantic 
segmentation tests, such as in \cite{BAD2015,LON2015}. 

In the following, first we describe the way how we embed 
our spatio-temporal module into FCNs and dilation convolution networks. 
Then we describe the metrics used in the evaluation process. 
After that we report our experiments on CamVid and NYUDv2. 
Finally, we discuss the performance of our method. 

\subsection{Embedding the Spatio-Temporal Module in FCN Networks}
\label{sec:E_FCN}

FCN-8 and FCN-32 \cite{LON2015} are fully convolutional versions 
of VGG-16 with some modifications to combine features of shallow layers 
with more precise spatial information with features of
deeper layers which have more precise semantic information.

As mentioned in Section~\ref{sec:pm}, it is of benefit to 
embed the spatio-temporal module on top of the deepest layers. 
Thus we embed our spatio-temporal module on top of the 
$fc7$ layer of FCN-8 and FCN-32. The $fc7$ is the deepest 
fully convolutional layer which has large corresponding 
receptive fields in the input image. This layer extracts 
features which represent more semantic information in comparison 
to shallower layers. 

An example of this modification of an FCN-Alexnet is shown 
in Fig.~\ref{fig:1}. After embedding our spatio-temporal 
module in FCN-8 and FCN-32 networks, we call them 
STFCN-8 and STFCN-32. Our spatio-temporal module 
consists of LSTMs with 30 hidden nodes and 3 time-steps 
for the CamVid dataset. We fine-tuned our STFCN networks 
from pre-trained weights on PASCAL VOC \cite{Ever2011} 
provided by \cite{LON2015}. We used a 
momentum amount of 0.9, and a learning rate of 10e-5.

\subsection{Embedding Our Module in Dilated Convolution Networks}

A dilated convolution network is an FCN network which 
benefits from some modifications such as reducing down-sampling 
layers and using a context module which uses dilated convolutions. 
This module brings multi-scale ability to the network \cite{YU2016}.

The dilated8 network \cite{YU2016} consists of 
two modules, front-end and context. The front-end module 
is based on a VGG-16 network with some modifications. 
The context layer is connected on top of this module. 
The $fc7$ layer of the front-end layer provides the main 
spatial features with 4,096 maps. This network has an input 
of size $900 \times 1,100$. Because of removing some 
of its down-sampling layers, the $fc7$ layer has an output of 
size $66 \times 91$ which defines a high dimension for 
spatio-temporal computations. For overcoming this complexity problem, 
we down-sampled the output of this layer by a convolution 
layer to the size of $21 \times 30$, and fed it to our 
spatio-temporal module. Then, the spatio-temporal features 
are fed to a convolutional layer to decrease their maps 
to the size of the {\it final} layer of the front-end module. 

After resizing the maps, features are fed to a deconvolution 
layer to up-sample them to the size of the {\it final} layer 
output ($66 \times 91$). Finally, we fuse them with the front-end {\it final} 
layer by an element-wise sum operation over all features. 

\begin{figure*}[t]
	\center
	\includegraphics[width=1\linewidth]{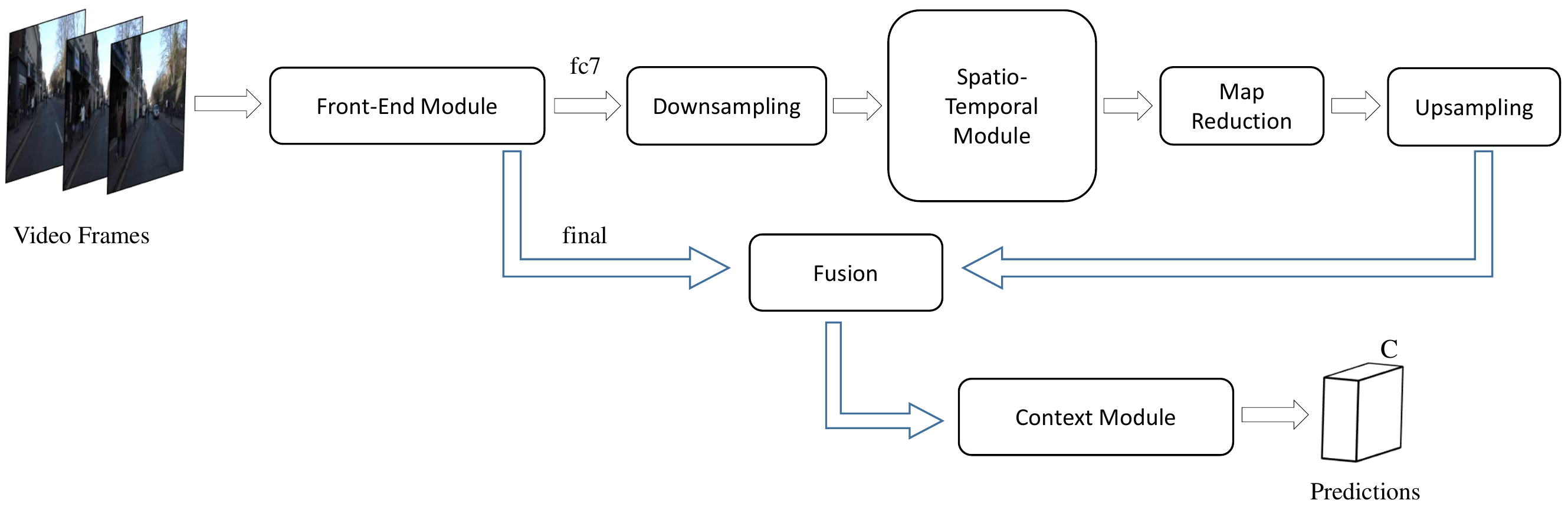}
	\caption
	{
		Our STDilation8 model architecture
	}
	\label{fig:STDilation8}
	\vspace{-3mm}
\end{figure*}

The fused features are fed to the context module. 
Let STDilated8 be the modified version of dilated8; see 
Fig.~\ref{fig:STDilation8}. The spatio-temporal module 
of STDilated8 consists of 30 hidden nodes of LSTMs with 
a time-step of 3. For training this network, we fixed the 
front-end module and fine-tuned the spatio-temporal and 
context modules with dilation8 pre-trained weights on 
CamVid. We used a momentum amount of 0.9, and 
a learning rate of 10e-5.

For better performance of the spatio-temporal module, we down-sampled the 
output of the $fc7$ layer of the dilation8 front-end module and fed it to 
the spatio-temporal module. Then we reduced the 
feature maps by a fully convolutional layer for a better 
description of the spatio-temporal features 
and make them the same size as the final layer of the front-end module. 
Finally we up-sample and fuse the spatio-temporal features 
with the final layer output and feed them into the context module.

\subsection{Quality Measures for Evaluation}

There are already various measures available for 
evaluating the accuracy of semantic segmentation. We describe most
commonly used measures for accuracy evaluation which we 
have used to evaluate our method.

\subsubsection{Mean intersection over union.}
Mean IU is a segmentation performance measure 
that quantifies the overlap of two objects by calculating the 
ratio of the area of intersection to the area of 
unions \cite{KLE2004,ZHA2015}. 
This is a popular measure since it penalizes both 
over-segmentation and under-segmentation 
separately \cite{SHA2015}. It is defined as follows:
\begin{eqnarray}
\frac{1}{n_{cl}}\cdot \sum_i {\frac{n_{ii}}{t_i + \sum_j {n_{ji}} - n_{ii}}}
\label{eq:iu}
\end{eqnarray}
where $n_{ii}$ is the number of pixels of class $i$ that is 
predicted correctly as belonging to class $i$, $t_i$ 
is the total number of pixels in class $i$, and 
$n_{cl}$ is the number of classes.

\subsection{CamVid}
\label{sub:camvid}

The {\it Cambridge-driving labelled video database} (CamVid) \cite{BRO2008} 
is a collection of videos with object-class semantic labels, 
complete with meta-data. The database provides ground truth 
labels that associate each pixel with one of 32 semantic classes. 
Like in \cite{STU2012}, we partitioned the dataset into 367 
training images, 100 validation, and 233 test images. 
Eleven semantic classes are used in the selected images.

For FCN-8, FCN-32, STFCN-8, and STFCN-32,
the images are down-sampled to $400 \times 400$. 
For dilation8 and STDilation8, the images are down-sampled 
to $640 \times 480$. As mentioned before, we used time-step 3 
for our spatio-temporal module which means that we feed a 
sequence of 3 frames to our spatio-temporal networks. 

The reason for choosing number 3 is that the annotated frames 
of CamVid have a distance of 30 frames to each other. In fact when 
we use 3 frames as a sequence, the first and last frame of the sequence 
have a distance of 90 frames. Using more annotated frames is 
computationally possible because of the given LSTM abilities, 
but it is semantically wrong because of the high amount of 
changes in the frames. 

Our results of FCNs and STFCNs tests 
on CamVid are shown in Table~\ref{tab:1}. It appears that 
adding our spatio-temporal module into FCN networks 
shows an improvement of their performance by close to one percent.
Results for dilation8 and STDilation8 tests on CamVid are 
shown in Table~\ref{tab:2}. The effect of the spatio-temporal module
is here an improvement by $0.8 \%$. Improvements are in both cases 
not ``dramatic'' but consistent. Note that reports about improvements
in the semantic segmentation area are typically in the sub-one-percent
range \cite{KUN2016,YU2016,LON2015}.

\begin{table}[t]
	\caption{Evaluating FCNs and STFCNs for video semantic segmentation on Camvid
	    (i.e. without or with our spatio-temporal module)}
	\begin{center}
		\begin{tabular}{lccccc}
			\hline
			   & \quad FCN-32s\qquad & \qquad STFCN-32s\qquad & \qquad FCN-8s\qquad & \qquad STFCN-8s\\
			\hline
			Mean IU & \quad $46.1\%$ & \quad $46.9\%$ & \quad\quad $49.7\%$ & \quad $50.6\%$  \\
			\hline
		\end{tabular}
	\end{center}
	\label{tab:1}
\end{table}

\begin{table}[t]
	\vspace{-3mm}
	\caption{Evaluating dilated convolution 
	    networks, without or with our module on Camvid}
	\begin{center}
		\begin{tabular}{lccccc}
			\hline
			& \qquad Dilation8\qquad  & \qquad STDilation8 (90 Frames) \qquad $ $\\
			\hline
			Mean IU & \qquad$65.3\%$  & \qquad$65.9\%$\\
			\hline
		\end{tabular}
	\end{center}
	\label{tab:2}
	\vspace{-6mm}
\end{table}

Dilation8 achieves the best results in comparison to other work, and this is
due to the power of multi-scale semantic segmentation. 
STDilation8 achieves even slightly better results because of 
benefits from temporal features. Detailed results on the CamVid test 
set are reported in Table~\ref{tab:tbl2}. Our model outperforms prior 
state-of-the-art work.

\begin{table}[t]
	\caption{Our STDilation8 improves Dilation8 and outperforms prior work on Camvid}
	\label{tab:tbl2}       
	\begin{tabular}{l||l|l|l|l|l|l|l|l|l|l|l||l}
		\hline\noalign{\smallskip}
		& \multicolumn{1}{c}{\begin{turn}{90}Building\end{turn}} & \multicolumn{1}{c}{\begin{turn}{90}Tree\end{turn}} & \multicolumn{1}{c}{\begin{turn}{90}Sky\end{turn}} & \multicolumn{1}{c}{\begin{turn}{90}Car\end{turn}} & \multicolumn{1}{c}{\begin{turn}{90}Sign\end{turn}} & \multicolumn{1}{c}{\begin{turn}{90}Road\end{turn}} & \multicolumn{1}{c}{\begin{turn}{90}Pedestrian\end{turn}} & \multicolumn{1}{c}{\begin{turn}{90}Fence\end{turn}} & \multicolumn{1}{c}{\begin{turn}{90}Pole\end{turn}} & \multicolumn{1}{c}{\begin{turn}{90}Sidewalk\end{turn}} & \multicolumn{1}{c}{\begin{turn}{90}bicyclist\end{turn}} & \multicolumn{1}{c}{\begin{turn}{90}mean IU\end{turn}}\\
		\noalign{\smallskip}\hline\noalign{\smallskip}
		ALE \cite{LAD2009} & 73.4 & 70.2 & 91.1 & 64.2 & 24.4 & 91.1 & 29.1 & 31.0 & 13.6 & 72.4 & 28.6 & 53.6 \\
		SuperParsing \cite{TIG2010} & 70.4 & 54.8 & 83.5 & 43.3 & 25.4 & 83.4 & 11.6 & 18.3 & 5.2 & 57.4 & 8.90 & 42.0 \\
		Liu and He \cite{LIU2015} & 66.8 & 66.6 & 90.1 & 62.9 & 21.4 & 85.8 & 28.0 & 17.8 & 8.3 & 63.5 & 8.50 & 47.2 \\
		SegNet \cite{BAD2015} & 68.7 & 52.0 & 87.0 & 58.5 & 13.4 & 86.2 & 25.3 & 17.9 & 16.0 & 60.5 & 24.8 & 46.4 \\
		STFCN-8 & 73.5 & 56.4 & 90.7 & 63.3 & 17.9 & 90.1 & 31.4 & 21.7 & 18.2 & 64.9 & 29.3 & 50.6 \\
		DeepLab-LFOV \cite{CHE2015} & 81.5 & 74.6 & 89.0 & 82.2 & 42.3 & 92.2 & 48.4 & 27.2 & 14.3 & 75.4 & 50.1 & 61.6 \\
		Dilation8 \cite{YU2016} & 82.6 & 76.2 & 89.9 & 84.0 & 46.9 & 92.2 & 56.3 & 35.8 & 23.4 & 75.3 & 55.5 & 65.3 \\
		Dilation + FSO \cite{KUN2016} & 84.0 & 77.2 & 91.3 & 85.6 & 49.9 & 92.5 & 59.1 & 37.6 & 16.9 & 76.0 & 57.2 & 66.1 \\
	    STDilation8 & 83.4 & 76.5 & 90.4 & 84.6 & 50.4 & 92.4 & 56.7 & 36.3 & 22.9 & 75.7 & 56.1 & 65.9 \\
		\noalign{\smallskip}\hline
	\end{tabular}
	\vspace{-3mm}
\end{table}

\begin{figure*}[t]
	\center
	\includegraphics[width=0.9\linewidth]{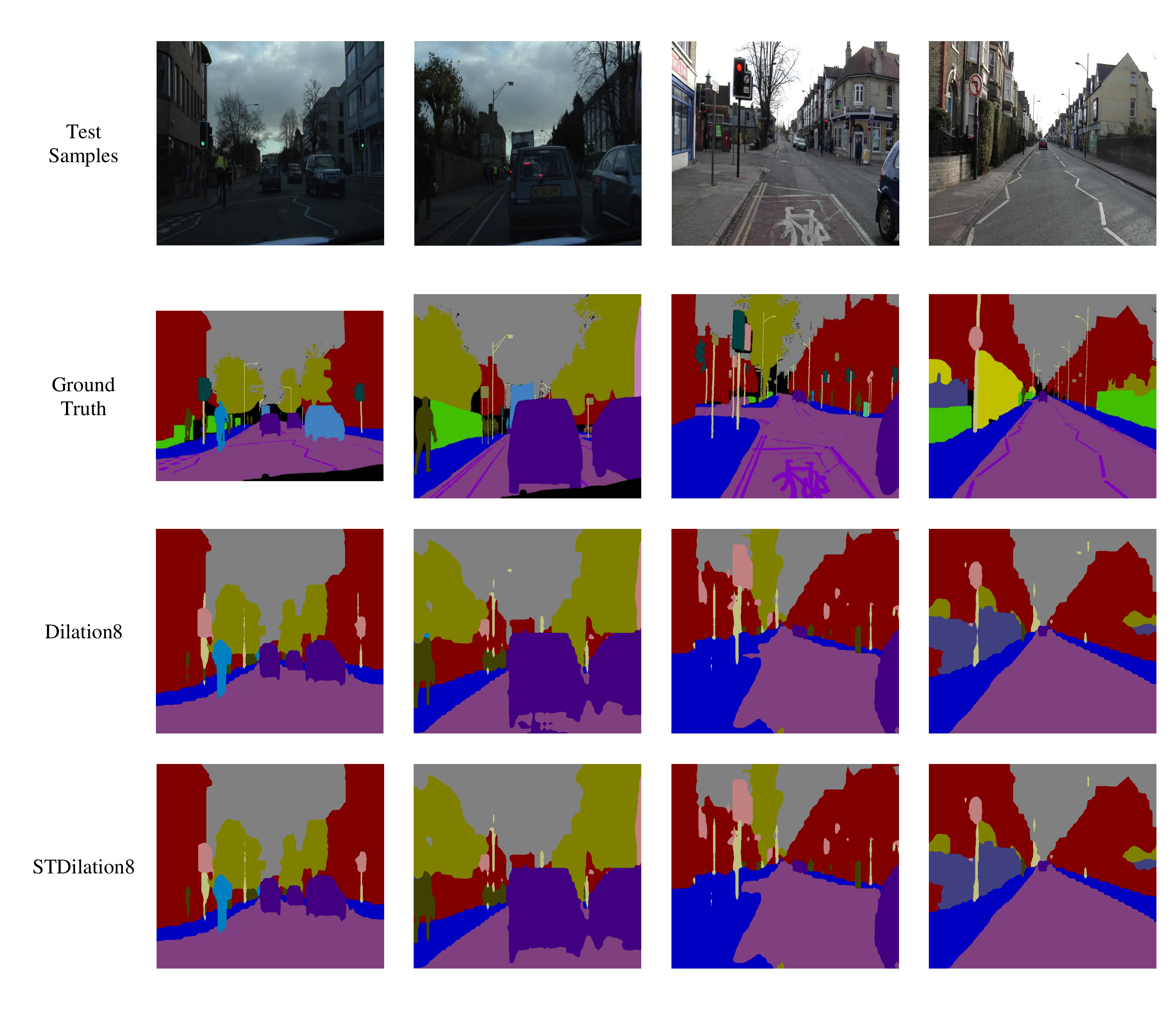}
	\caption
	{
		Outputs on CamVid. {\it Top to bottom rows:} Test samples, 
		ground truth, Dilation8 \cite{YU2016}, and STDilation8
	}
	\label{fig:Capture}
	\vspace{-6mm}
\end{figure*}

Table~\ref{tab:tbl2} shows that some approaches are competitive to related work 
such as Liu and He \cite{LIU2015} with a performance superiority by 0.8 percent 
compared to SegNet \cite{BAD2015}. In contrast, other approaches with a 
new base architecture achieved a better performance. Since our approach is 
based on FCN \cite{LON2015} or Dilation8 \cite{YU2016} methodologies, with 
our introduced spatio-temporal module, performance enhancement is close to one 
percent on FCN network, and close to 0.8 percent on Dilation8 architecture;
both can be considered as a being a noticable enhancement. Dilation + FSO \cite{KUN2016} has been published recently based on Dilation8 architecture and became state-of-the-art video semantic segmentation method. Our approach differs from FSO in several ways:
\begin{itemize}
\item Our approach does not need any pre-processing or 
   feature optimization for result enhancement. In contrast, FSO has used optical flow as a feature set to be used by a CRF model. This is a computational operation which  can be considered as a weakness for a semantic segmentation method. Computation efficiency and speed is  very crucial in some tasks, such as, online video processing in advanced driver assistance systems. Some researches are ongoing to resolve optical flow computational cost by using convolutional networks \cite{fischer2015}. 
\item We used time-step 3 for our spatio-temporal module to use the CamVid dataset annotations as-is without any preprocessing. This simplicity in design and configuration, is one of the strengths of our work.
\item Our approach proposes an end-to-end network for 
    semantic video segmentation which consists of
    spatial and temporal features altogether.
\item In our approach we proposed a neural network based module for 
    transforming traditional, fully convolutional 
    networks into spatio-temporal CNNs. It can 
    also be used for other related video processing tasks.
\end{itemize}

Also, we embedded our spatio-temporal module into FCN-Alexnet 
and evaluated its performance with and without our spatio-temporal module. 
Our spatio-temporal module improved its performance on CamVid dataset. 
Because the basic FCN-Alexnet has a low performance for semantic 
segmentation as described in \cite{LON2015}, so we decided not to include
details into this paper.

\subsection{NYUDv2}
\label{sub:nyudv2}

The NYU-Depth V2 data set is comprised of 
video sequences from a variety of indoor scenes 
recorded with an RGB-Depth camera \cite{SIL2012}. It features 
1,449 densely labelled pairs of aligned RGB and depth images, including
464 new scenes taken at three cities, and 407,024 new 
unlabeled frames. 

We selected this dataset to evaluate the effect of 
multi-modal learning on our spatio-temporal module. Also, we tested our 
method on two totally different datasets (outdoor vs. indoor) to 
evaluate its flexibility.
One problem of this dataset is that its annotated frames vary in 
length of sequences per subject or location. 
Thus, for this dataset, we do not use a constant time step 
for the spatio-temporal module. We fed sequences of 
different lengths based on their location. This problem showed 
its effect on results by decreasing the amount of improvements compared to
none-temporal models.

Gupta et al. \cite{GUP2014} coalesced NYU-Depth V2 into 40 classes. 
Similar to \cite{LON2015} we report results on a standard split into 
795 training images and 654 test images. We selected our models 
based on \cite{LON2015} to be able to evaluate an embedding 
of our spatio-temporal module into their models. We use 
FCN-32s RGB and FCN-32s RGBD models to embed our 
spatio-temporal module in the way as explained before. 
Tests on NYUDv2 data are reported in Table~\ref{tab:4}. 

\begin{table}[t]
	\vspace{-3mm}
	\caption{
	Evaluating STFCNs, FCNs, and dilated convolution networks for 
	    semantic video segmentation on NYUDv2
	    }
	\begin{center}
		\begin{tabular}{lccccc}
			\hline
			Network   & Pixel Accuracy\qquad & \qquad Mean Accuracy\qquad & \qquad Mean IU  \\
			\hline
			Gupta \textit{et al.} \cite{GUP2013} & $60.3\%$ & $-$ & $28.6\%$  \\
			FCN-32s RGB \cite{LON2015} &60.0\% & 42.2\% & 29.2\%\\
			FCN-32s RGBD \cite{LON2015} & 61.5\% & 42.4\% & 30.5\%\\
			STFCN-32s RGB &60.9\% & 42.3\% & 29.5\%\\
			STFCN-32s RGBD & 62.1\% & 42.6\% & 30.9\%\\
			\hline
		\end{tabular}
	\end{center}
	\label{tab:4}
	\vspace{-6mm}
\end{table}

Results show in this case the enhancement effect of the spatio-temporal module on 
FCN-32s RGB and FCN-32s RGBD compared to the related networks 
FCN-32s RGB and FCN-32s RGBD, respectively.

\subsection{Discussion}

We showed the power of our spatio-temporal module by 
embedding it into other known spatial, fully convolutional networks. 
In fact we introduced a spatio-temporal, fully convolutional network 
for extracting spatio-temporal features from video data 
and evaluated it based on two semantic segmentation case studies. 

Our module benefits from the LSTM characteristics 
and is able to handle long-short term sequences. In our tests 
we were only able to use a limited number of video frames 
as being one sequence 
because of the limited number of available annotated frames. 
The method should also be tested on datasets with more extensive 
sets of annotated frames to check the effect of sequence length 
on the performance of the system. 
It is possibly also of value to check the effect of involving 
unannotated frames into input sequences by using prior or 
posterior annotated frames in the system.

\section{Conclusions}
\label{sec:con}

In this paper we proposed a new architecture 
for spatio-temporal feature extraction from video. 
We designed and used this architecture for 
semantic video segmentation.  
First, a pre-trained CNN model was turned into an 
FCN model by changing classification layers into fully 
convolutional layers. In this phase, spatial features from 
input frames can be used for classification. But, in 
semantic video segmentation, relationships between frames 
can provide very useful information and enhance 
the accuracy of the segmentation program. Therefore, 
LSTM modules have been used to take advantage
of temporal features.
This architecture has been proposed as an end-to-end 
trainable model and can also be used for other vision tasks. 
Also, it does not need to be a pre-processing or post-processing 
module only, as we have seen in some other approaches. 

We illustrated the performance of our architecture by 
embedding our spatio-temporal module into some 
state-of-the-art fully convolutional networks, such as FCN-VGG, 
and dilation convolution. Other types of LSTM modules have been 
proposed recently and have shown promising results for some 
vision tasks. Applying these newly proposed modules 
may enhance further the architecture of our spatio-temporal module,
e.g. for scene understanding, anomaly detection in video, 
video captioning, object tracking, activity recognition, and so forth.

\newpage

\end{document}